\documentclass[pdflatex,sn-mathphys-num]{sn-jnl}% Math and Physical Sciences Numbered Reference Style
\usepackage{graphicx}%
\usepackage{multirow}%
\usepackage{amsmath,amssymb,amsfonts}%
\usepackage{amsthm}%
\usepackage{mathrsfs}%
\usepackage[title]{appendix}%
\usepackage{xcolor}%
\usepackage{textcomp}%
\usepackage{manyfoot}%
\usepackage{booktabs}%
\usepackage{algorithm}%
\usepackage{algorithmicx}%
\usepackage{algpseudocode}%
\usepackage{listings}%
\theoremstyle{thmstyleone}%
\theoremstyle{thmstyletwo}%

\theoremstyle{thmstylethree}%

\begin{document}

\title[CNN-based classifier for subject-independent MBTI Extraversion/Introversion preference classification from short-time EEG signals]{CNN-based classifier for subject-independent MBTI Extraversion/Introversion preference classification from short-time EEG signals}


\author*[1]{\fnm{Filip} \sur{Postepski}}\email{filip.postepski@mail.umcs.pl}

\author[1]{\fnm{Grzegorz M.} \sur{Wojcik}}\email{gmwojcik@umcs.edu.pl}
%%\equalcont{These authors contributed equally to this work.}

\affil*[1]{%
\orgdiv{Department of Neuroinformatics and Biomedical Engineering, Institute of Computer Science and Mathematics},
\orgname{Maria Curie-Skłodowska University},
\orgaddress{
\street{Akademicka 9},
\city{Lublin},
\postcode{20-031},
\state{Lubelskie},
\country{Poland}
}}


\abstract{This study examines whether short, minimally preprocessed EEG segments contain sufficient information to distinguish between individuals classified as extraverted or introverted according to the Myers--Briggs Type Indicator (MBTI). EEG recordings from 50 participants were divided into 4-s segments and analyzed using a subject-independent nested 5-fold cross-validation procedure to compare EEGNet, EEGProgress, and a newly proposed compact convolutional architecture, SSCNet. SSCNet achieved the highest mean accuracy (0.6481), AUC (0.7063), and F1-score (0.6276). The proposed architecture also showed the lowest variability across folds. SSCNet demonstrated the best computational efficiency, with the shortest median training epoch time and the highest training throughput despite having a similar number of trainable parameters to EEGProgress. The results suggest that MBTI Extraversion--Introversion preferences may be moderately predictable from short time-domain EEG segments without manual spectral feature extraction, while highlighting the importance of subject-independent validation and computationally efficient model design.}




\maketitle

\section{Introduction}\label{sec1}

Individual psychological characteristics are commonly assessed using psychological questionnaires. Such instruments provide standardized measures of personality traits, cognitive styles, emotional characteristics, and other individual differences. However, questionnaire-based assessment relies on self-reported information and therefore reflects how individuals describe or perceive themselves. The growing field of personality neuroscience has introduced a complementary perspective by investigating whether individual psychological characteristics are also reflected in measurable patterns of brain activity\cite{deYoung2022Personality}.

Electroencephalography (EEG) is particularly suitable for this type of research because it is a non-invasive brain activity measurement. It records electrical activity from the scalp with high temporal resolution and at relatively low cost compared to other neuroimaging modalities, such as fMRI or PET. It can be also less stressful for patients.\cite{murphy1997adult, eeg_signal_processing}

Extraversion and introvesion are one of the key personality dichotomies covered by Myers-Briggs Type Indicator(MBTI) questionnaire, alongside Sensing-Intuition ("S-I"), Thinking-Feeling ("T-F") and Judging-Perceiving ("J-P"). The questionnaire distinguishes between introversion and extraversion according to how individuals engage with their environment. People with an introverted preference ("I") are more likely to focus on their internal world and often prefer reflection, contemplation, and activities performed independently. On the other hand, those with an extraverted preference ("E") tend to seek stimulation from the external world and are more likely to feel energized by communication, social activity, and interaction with others. Although the MBTI remains widely used, its dichotomous representation of 
personality preferences and some of its psychometric properties have been 
criticized in the literature \cite{king2020myers}.

A series of studies by Bhardwaj et al. have demonstrated that classification of personality dimensions collected from MBTI questionnaire is possible with high accuracy using genetic programming\cite{bhardwaj2022personality, bhardwaj2022portable}. However in this cycle of papers fast-fouries tranform (FFT) was used for signal features extraction.

Researchers have investigated whether extraversion-introversion differences are reflected in EEG recordings. However, evidence for stable electro-physiological markers of extraversion remains 
inconsistent. A meta-analysis of 79 independent samples comprising 5700 
participants found that resting frontal EEG asymmetry explained less than 
0.4\% of the variance in extraversion and neuroticism 
\cite{kuper2019resting}. More recent studies have also produced mixed results. 
Pacheco et al. reported that Big Five personality traits could be decoded from 
periodic resting-state EEG power with modest predictive correlations of 
approximately $r=0.20$ \cite{PACHECO2024}, whereas a preregistered 
replication involving 772 participants did not find meaningful prediction of 
Big Five traits from resting-state spectral EEG power 
\cite{FRUEHLINGER2024}. Although the Big Five framework was used in those studies rather than the MBTI, their results indicate that EEG--personality associations may be weak and sensitive to the 
signal representation and analysis pipeline.

Machine-learning studies have provided evidence that personality related-information can be extracted from EEG under some specific conditions. Baumgartl et al. achieved a balanced accuracy of 60.6\% when distinguishing extraversion from introversion using resting-state EEG and a 
random forest classifier \cite{baumgartl2020measuring}. More recently, Penava 
and Buettner proposed a subject-independent CNN for predicting all five Big 
Five personality traits from resting-state EEG \cite{penava2025novel}. However, 
their pipeline first transformed EEG recordings into fine-grained spectral 
power features, removing the temporal domain before classification. Therefore, 
it remains unclear whether the Extraversion--Introversion preference can be 
classified directly from short, minimally preprocessed time-domain EEG segments 
without manual spectral feature extraction, and whether CNN-based architectures 
can generalize across individuals. This constitutes the main research problem 
addressed in the present study.

Key contributions of this study are as follows:

\begin{itemize}
    \item Find out if any 4-s EEG segment can be classified as signal taken from Extraverted or Introvered person using CNN-based classifiers according to MBTI. 
    \item Compare classification performance with two existing architectures of CNN-based models for EEG signal: EEGNet and EEGProgress under identical and evaluation conditions. 
    \item Propose and evaluate a new CNN-based architecture, with the emphasis on computational efficiency, predictive performance, and robustness across subject-independent cross-validation folds
\end{itemize}
% The Introduction section, of referenced text \cite{bib1} expands on the background of the work (some overlap with the Abstract is acceptable). The introduction should not include subheadings.

% Springer Nature does not impose a strict layout as standard however authors are advised to check the individual requirements for the journal they are planning to submit to as there may be journal-level preferences. When preparing your text please also be aware that some stylistic choices are not supported in full text XML (publication version), including coloured font. These will not be replicated in the typeset article if it is accepted. 

\section{Materials and methods}

All models were trained on the same dataset. The dataset consisted of 50 recordings of EEG signal recorded for 20 minutes. Participants were listening to guided imagery recording (25 subjects) or resolving mental tasks (also 25 subjects). Also half of a dataset were represented by extraverts and the other half was represented by introverts. The Extraversion--Introversion labels
were also approximately balanced across the two experimental conditions, as shown in Table~\ref{tab:group_personality_distribution}.
\begin{table}[ht]
\centering
\caption{Distribution of Extraversion--Introversion labels across the
experimental conditions.}
\label{tab:group_personality_distribution}
\begin{tabular}{lccc}
\hline
\textbf{Condition} & \textbf{Introversion} & \textbf{Extraversion} & \textbf{Total} \\
\hline
Guided Imagery & 13 & 12 & 25 \\
Mental Task    & 12 & 13 & 25 \\
\hline
Total          & 25 & 25 & 50 \\
\hline
\end{tabular}
\end{table}

\subsection{Participants and ethical approval}

The EEG data used in the present study were acquired as part of a previously established experimental protocol. During the experiment, participants underwent either a Guided Imagery session intended to induce relaxation or a Mental Workload condition involving cognitively demanding memory-recall tasks. EEG recordings were collected under controlled laboratory conditions. The experimental procedure, EEG acquisition setup, and characteristics of the investigated conditions have been described in previous studies. The same EEG recordings have previously been investigated primarily in the context of distinguishing experimentally induced mental states\cite{postepski2025recurrent}. This time the research sample is bigger. The present analysis included EEG recordings from 50 anonymized participants recruited according to the same eligibility criteria as in our previous study, comprising healthy, right-handed young adults aged 19–24 years. Before the EEG recording the study participants provided written informed consent and completed the Myers–Briggs Type Indicator (MBTI) personality questionnaire. The study protocol was reviewed and approved by the Maria Curie-Sklodowska Bioethics Committee. All methods were performed in accordance with the relevant guidelines and regulations.

\subsection{Experimental procedure and EEG acquisition}
For this study EGI 256 dense-array was used. For the signal recording Net Station 4.5.4 was used. The participants were ask to lay down with eyes closed to reduce artifacts from eye blinks and movement. The recordings were 20 minutes long. EEG signal was collected with sampling rate set to 250 Hz. The EGI GSN 130 dense-array cap was used for signal acquisition.

\subsection{EEG signal preprocessing and dataset preparation}
For this study 50 signals recorded from extraverted and introverted subjects were selected. The signal was preprocessed using mne python toolkit. Minimal preprocessing pipeline was done as mentioned it our previous work\cite{postepski2025recurrent}. Bad channels were removed and interpolated using RANSAC algorithm. Each signal was filtered using 1 - 45 Hz band-pass filter. Next the whole recordings were split into 4-s segments. Only segments where subjects were listening to Guided Imagery recording or were solving mental-tasks were selected for classification. From the whole recording segments where in mental tasks recording when participats were listening to instructions were removed. Accordingly, complementary segments in the guided imagery recordings were also removed. This resulted in total of 122 segments for each subject. From all of 256 electrodes 69 were left. The selection was backed by the reduction table from full cap to 10-10 system in the \cite{geodesics2003net} Geodesics Sensor Net manual.

\subsection{Subject-independent cross-validation}
\begin{figure}
    \centering
    \includegraphics[width=1\linewidth]{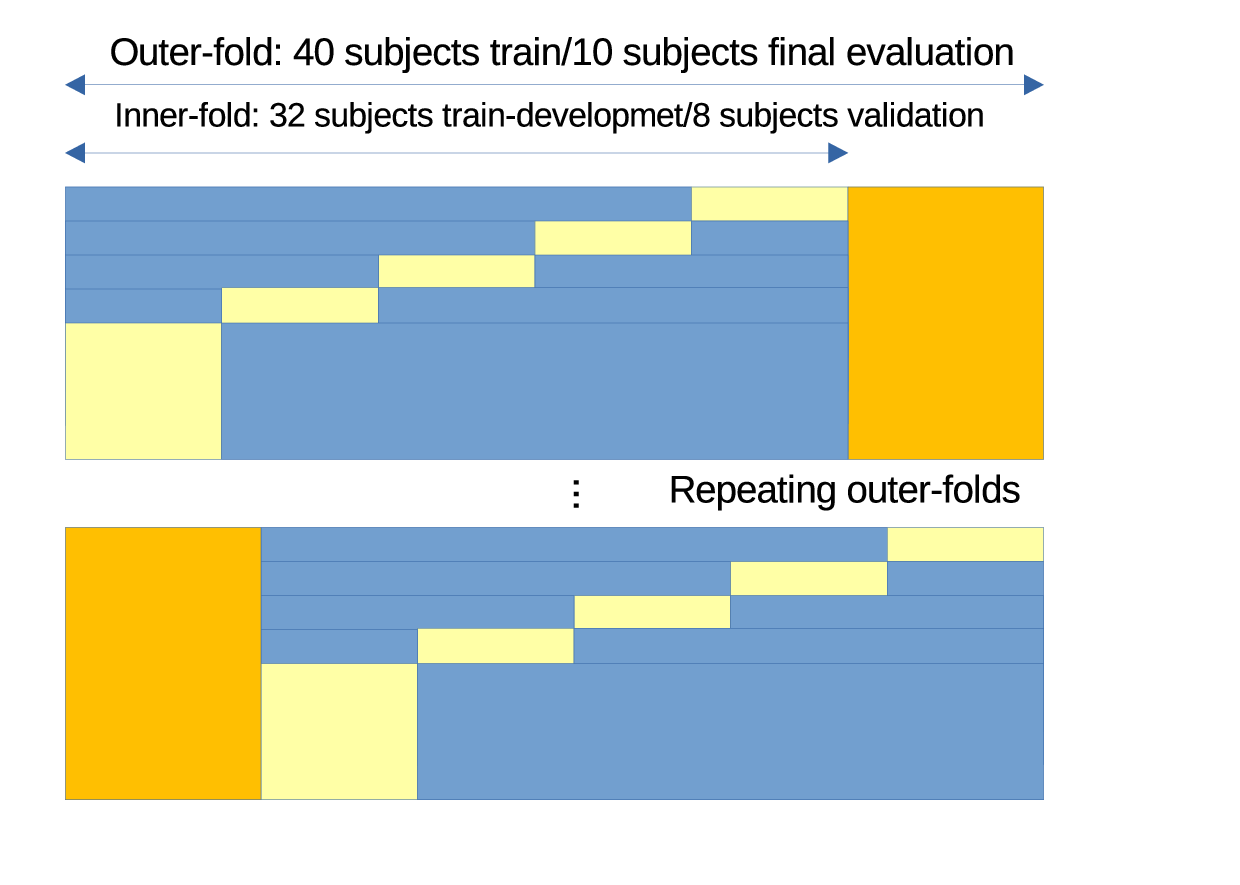}
    \caption{Nested cross-validation procedure used for models development and evaluation.}
    \label{fig:nested_corss-validation}
\end{figure}

For this study subject-independent nested 5-folds cross-validation procedure was applied. 40 subjects in each fold were used for development and hyperparameters selection. 10 subjects were left from each fold for final evaluation of 3 models generalization performance. Within those 40 subjects the training set contained 32 training subjects and 8 validation subjects. All EEG windows originating from a given subject were assigned exclusively to one partition, preventing subject-level data leakage. Identical cross-validation splits were used for all three architectures to ensure a direct and fair comparison. The nested 5-fold cross-validation procedure is depicted in Figure ~\ref{fig:nested_corss-validation}. The inner-folds validation datasets and outer-folds test datasets were also balanced according to the target label and stratified with respect to the recording condition.

\subsection{CNN-based classification models}
Two CNN-based models were selected to compare with proposed architecture. EEGNet was used as a benchmark architecture. We also selected EEGProgress for comparison as it proposed  newer different approach to EEG signal classification.

\subsubsection{EEGNet}
EEGNet was selected as one of the reference architectures for EEG signal 
classification. The model is a compact convolutional neural network originally 
proposed by Lawhern et al. for EEG-based brain--computer interface applications. 
Its architecture combines temporal convolutions with depthwise spatial filtering 
and separable convolutions, enabling the extraction of temporal and spatial 
features while maintaining a relatively low number of trainable parameters\cite{Lawhern_2018_eegnet}.

The implementation used in this study was based on the original EEGNet 
architecture and was adapted for binary classification. The number of temporal 
filters $F1$ was set to 8, the number of pointwise filters $F2$ to 16, and the 
depth multiplier $D$ to 2. The temporal kernel length was set to 125 samples. 
The original output layer was replaced with a single-neuron layer using the 
Sigmoid activation function to provide the probability of belonging to one of 
the two personality classes. The configuration was also used in previous research based on smaller dataset\cite{postepski2025recurrent}. The resulting model contained 3,397 trainable parameters.

\subsubsection{EEGProgress}
EEGProgress was selected as the second reference architecture. It was described as fast and lightweight architecture. It introduces different architecture idea which is to extract progressively information from electrodes, brain regions and electrodes\cite{CHEN2024107901}. 

The original EEGProgress framework combines a topological permutation stage
with a progressive convolutional feature extractor composed of prior, electrode,
region, and hemisphere convolution blocks. This design replaces a single global
spatial convolution with several successive convolutional operations, reducing
model complexity while preserving multi-scale spatial information. In the
implementation used in the present study, EEGProgress was adapted to the
69-channel EEG input and binary Extraversion--Introversion classification task.
The resulting architecture contained 16,245 trainable parameters.

\subsubsection{SSCNet}
SSCNet was proposed as a compact CNN-based architecture designed to process
short multichannel EEG segments while limiting computational complexity. The
model receives 4-s EEG windows represented by 1000 time samples and 69 channels.
The first layer is a separable one-dimensional convolution with 64 filters, a
kernel length of 90 samples, and a stride of 5. At the sampling frequency of
250~Hz, the kernel covers approximately 360~ms of the original EEG signal.
A relatively long temporal kernel was selected to enable the extraction of
temporal patterns extending over several hundred milliseconds, while the stride
simultaneously reduces the temporal resolution and computational cost of the
subsequent layers. Separable convolution was employed to reduce the number of
trainable parameters compared with a conventional convolution while still
allowing temporal feature extraction and subsequent interaction between EEG
channels.

The second separable convolution uses 32 filters, a shorter kernel of 9, and a
stride of 4. At this stage, the convolution operates on the feature
representation generated by the first layer and further reduces the temporal
dimension while extracting more local combinations of previously detected
patterns. Considering both convolutional layers, each resulting feature covers
approximately 130 samples, corresponding to about 520~ms of the original
recording. Spatial dropout with a rate of 0.05 was applied after both separable
convolutional layers to reduce dependence on individual feature maps while
preserving the temporal structure of the signal. The number of filters is
progressively reduced from 64 to 32 and subsequently to 16, forming a compact
feature bottleneck.

\begin{figure}
    \centering
    \includegraphics[width=1\linewidth]{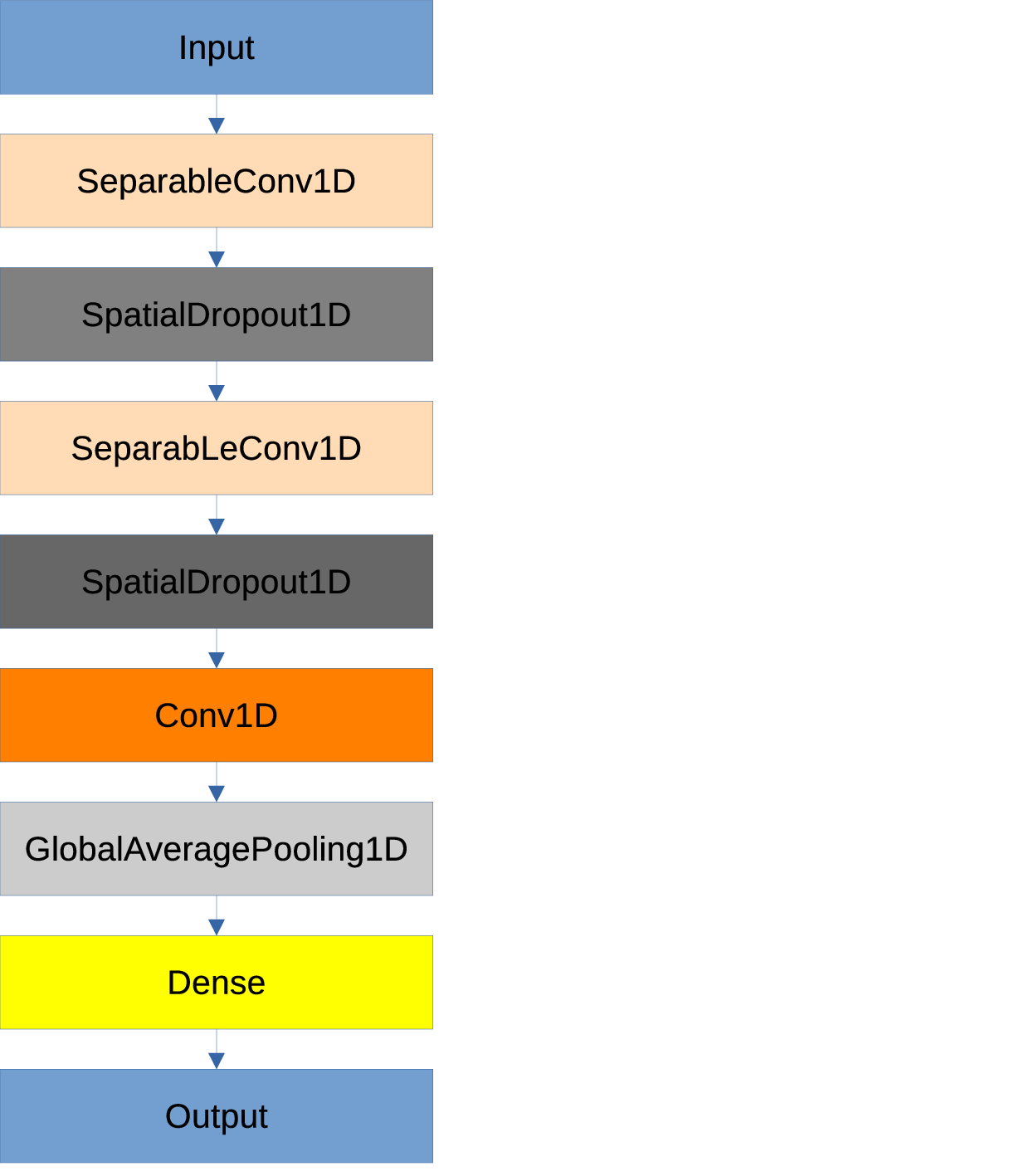}
    \caption{SSCNet model architecture.}
    \label{fig:sscnet_architecture}
\end{figure}

A $1 \times 1$ convolution with 16 filters was then used to combine and compress
the extracted feature representation without further reducing its temporal
resolution. Instead of flattening the convolutional output and introducing
additional fully connected layers, Global Average Pooling was applied to
aggregate the learned features over the complete EEG segment. This design
substantially limits the number of trainable parameters and makes the final
prediction less dependent on the exact temporal position of a particular
activation within the 4-s window. Finally, a single neuron with a Sigmoid
activation function was employed for binary Extraversion--Introversion
classification. The resulting SSCNet architecture contained 13,891 trainable
parameters. The architecture design is shown in Figure ~\ref{fig:sscnet_architecture}.

\subsection{Training procedure and hyperparameter selection}
Hyperparameter selection was performed exclusively within the inner cross-validation loop of each outer training partition. All of the models were checked of the outer folds only once with no further optimization for the fair comparison. Also each model in the final stage was retrained on the same fold using 3 different randomly selected initialization seeds. 
Each architecture was tested with the same optimizer Adam and the loss function was binary cross-entropy. The learning rates were tested from established set of values: 0.003, 0.001, 0.0003 and 0.0001. Each model during 5 inner-folds was trained for maximum number of 150 epochs, but early stopping was enabled with patience of 15 epochs to select the best averaged epoch. The loss parameter was monitored for the best epoch selection.

\subsection{Evaluation metrics}
The classification performance of the evaluated architectures was assessed using
accuracy (ACC), area under the receiver operating characteristic curve (AUC),
F1-score, and Matthews correlation coefficient (MCC). The use of multiple
evaluation measures provides a more comprehensive assessment of binary
classification performance than accuracy alone \cite{sokolova2009systematic}.

Accuracy represents the proportion of correctly classified samples among all
evaluated samples and is defined as:

\begin{equation}
ACC = \frac{TP + TN}{TP + TN + FP + FN},
\end{equation}

where $TP$, $TN$, $FP$, and $FN$ denote true positives, true negatives, false
positives, and false negatives, respectively.

The F1-score is the harmonic mean of precision and recall and provides a combined
measure of the classifier's ability to correctly identify positive samples while
limiting false-positive predictions \cite{sokolova2009systematic}. Precision and
recall are defined as:

\begin{equation}
Precision = \frac{TP}{TP + FP},
\end{equation}

\begin{equation}
Recall = \frac{TP}{TP + FN},
\end{equation}

and the F1-score is calculated as:

\begin{equation}
F1 = 2 \cdot
\frac{Precision \cdot Recall}{Precision + Recall}.
\end{equation}

The area under the receiver operating characteristic curve (AUC) was used as a
threshold-independent measure of the model's ability to discriminate between the
two classes \cite{fawcett2006roc}. The ROC curve represents the relationship
between the true-positive rate and false-positive rate across different
classification thresholds. AUC was calculated directly from the continuous
probability scores produced by the output layer rather than from thresholded
class predictions.

Matthews correlation coefficient was additionally employed as a balanced measure
of binary classification quality \cite{matthews1975comparison}. MCC takes all
four elements of the confusion matrix into account and is defined as:

\begin{equation}
MCC =
\frac{TP \cdot TN - FP \cdot FN}
{\sqrt{(TP+FP)(TP+FN)(TN+FP)(TN+FN)}}.
\end{equation}

For ACC, F1-score, and MCC, continuous model outputs were converted into binary
predictions using a classification threshold of 0.5. Performance metrics were
calculated independently for each random seed within each outer
subject-independent cross-validation fold. The three seed-specific results were
first averaged within each fold. Overall performance was subsequently reported
as the mean and standard deviation of the five resulting outer-fold values.

\subsection{Computational efficiency evaluation}
For the computational efficency evaluation we selected three measures: number of model parameters, median training epoch time and throughput which was number of training samples processed by the model during training per second. All of the models were trained for 20 epochs excluding the first which is typically longer due to the training initialization procedure. Then the results were averaged.

\section{Results}\label{sec2}

Taking into account ACC metric, EEGNet showed the greatest susceptibility to seed initialization. EEGProgress, despite showing less variability depending on the seed, achieved the worst average ACC scores. SSCNet outperformed EEGProgress in terms of average ACC across all folds after 3 initializations. It is also the most stable of all architectures in this regard. It also outperforms EEGNet when considering the average ACC scores in folds 3 and 1. The standard deviation of paired results across the test folds demonstrates SSCNet’s greatest resistance to random seed initialization.
The best fold across all networks for the three tested seeds is fold 4. EEGNet achieved the best average ACC in this fold after three initializations. SSCNet demonstrated the greatest robustness to initialization. EEGProgress showed the worst results. Each model achieved its worst results in a different fold. EEGProgress showed results below 0.5 in its worst fold.
Detailed results, aggregated as the average of 3 initialization trials with a random seed, are shown in the Table \ref{tab:acc_folds}.

\begin{table}[ht]
\centering
\caption{Classification accuracy (ACC) for the evaluated architectures across subject-independent cross-validation folds. Results are presented as mean $\pm$ standard deviation over three random seed initializations.}
\label{tab:acc_folds}
\begin{tabular}{lccc}
\hline
\textbf{Fold} & \textbf{EEGNet} & \textbf{EEGProgress} & \textbf{SSCNet} \\
\hline
Fold 0 & \textbf{0.6202 $\pm$ 0.0937} & 0.5672 $\pm$ 0.0815 & 0.5917 $\pm$ 0.0283 \\
Fold 1 & 0.5672 $\pm$ 0.1157 & 0.5190 $\pm$ 0.0304 & \textbf{0.6466 $\pm$ 0.0386} \\
Fold 2 & \textbf{0.6828 $\pm$ 0.0575} & 0.4725 $\pm$ 0.0513 & 0.6652 $\pm$ 0.0230 \\
Fold 3 & 0.5414 $\pm$ 0.1189 & 0.5750 $\pm$ 0.0590 & \textbf{0.6376 $\pm$ 0.0466} \\
Fold 4 & \textbf{0.7680 $\pm$ 0.1345} & 0.6156 $\pm$ 0.0766 & 0.6997 $\pm$ 0.0568 \\
\hline
\end{tabular}
\end{table}

SSCNet achieved the best overall performance in terms of ACC (0.6481), AUC (0.7063), and F1-score (0.6276), while also showing the lowest variability across subject-independent folds for these metrics. EEGNet obtained the highest MCC (0.2699), but its results were more variable across folds, particularly for AUC and F1-score. EEGProgress showed the weakest overall classification performance, with the lowest mean values of ACC, AUC, and MCC among the evaluated architectures. Results are shown in the Table \ref{tab:overall_performance}.

\begin{table}[ht]
\centering
\caption{Overall classification performance of the evaluated architectures. Results are presented as mean $\pm$ standard deviation across five subject-independent cross-validation folds, after averaging over three random seed initializations within each fold.}
\label{tab:overall_performance}
\begin{tabular}{lcccc}
\hline
\textbf{Model} & \textbf{ACC} & \textbf{AUC} & \textbf{F1-score} & \textbf{MCC} \\
\hline
EEGNet
& 0.6359 $\pm$ 0.0916
& 0.6943 $\pm$ 0.1098
& 0.6098 $\pm$ 0.1068
& \textbf{0.2699 $\pm$ 0.1583} \\

EEGProgress
& 0.5499 $\pm$ 0.0552
& 0.5786 $\pm$ 0.1162
& 0.5616 $\pm$ 0.0397
& 0.1157 $\pm$ 0.0924 \\

SSCNet
& \textbf{0.6481 $\pm$ 0.0395}
& \textbf{0.7063 $\pm$ 0.0513}
& \textbf{0.6276 $\pm$ 0.0444}
& 0.1928 $\pm$ 0.1217 \\
\hline
\end{tabular}
\end{table}

SSCNet demonstrated the highest computational efficiency, achieving the shortest epoch time (2.4511 s) and the highest throughput (3728.81 samples/s). Compared with EEGProgress, SSCNet was approximately 1.57 times faster per epoch and processed about 1.89 times more samples per second, despite having a comparable number of trainable parameters. EEGNet contained substantially fewer parameters, but exhibited the longest epoch time and the lowest throughput among the evaluated architectures. Table \ref{tab:computational_efficiency} shows the results of the several performance metrics from 20 epochs of training for each architecture tested.

\begin{table}[ht]
\centering
\caption{Computational efficiency of the evaluated architectures. Epoch time represents the average training time per epoch, while throughput denotes the number of training samples processed per second.}
\label{tab:computational_efficiency}
\begin{tabular}{lccc}
\hline
\textbf{Model} & \textbf{Parameters} & \textbf{Epoch time (s)} & \textbf{Throughput (samples/s)} \\
\hline
EEGNet      & \textbf{3,397}  & 10.2283 & 902.5927 \\
EEGProgress & 16,245 & 3.8402  & 1973.5584 \\
SSCNet      & 13,891 & \textbf{2.4511} & \textbf{3728.8085} \\
\hline
\end{tabular}
\end{table}

\section{Discussion}\label{sec3}

The results of this study indicate that short, 4-s EEG segments contain information that can be used to distinguish between individuals classified as introverted and extraverted according to the MBTI. Among the three evaluated CNN-based architectures, SSCNet achieved the highest mean ACC (0.6481), AUC (0.7063), and F1-score (0.6276). Moreover, SSCNet achieved lower variablility accross subject-idependent folds for AAC, AUC, and F1-score than the EEGNet and EEGProgress. These results suggest that the proposed architecture provides favorable balance between performance of prediction and stability over subjects groups of a 5-fold cross-validation. The obtained classification performance should be considered as moderate than as a highly accurate personality classifier.

The obtained results are rather consistent with previous studies suggesting that extraversion-related information can be extracted from EEG signals, although direct comparisons should be made cautiously because of differences in personality models, EEG acquisition protocols, signal representations, and evaluation procedures. 

An important finding of the presented study cover model stability. EEGNet was the most sensitive to random initialization, whereas SSCNet produced the lowest level of seed-related variability in four of five subject-independent folds. After averaging the three random initializations within each fold, SSCNet also exhibited the lowest variability across folds for three selected evaluation metrics such as: ACC ($0.6481 \pm 0.0395$), AUC ($0.7063 \pm 0.0513$), and F1-score ($0.6276 \pm 0.0444$). Those characteristics are relevant when it comes to EEG signal classification, where inter-subject variability can cause considerable differences in classifer performance over validation folds, as mentioned in latest research findings\cite{del2025role}.

Overall, the results show that classification of MBTI Extraversion-Introversion preferences from short EEG segments can be done at a moderate performance level. SSCNet achieved the best mean ACC, AUC and F1-score while at the same time showed lower variability accross subject-independent folds and greater computational efficency than the two reference architectures. The main advantage should therefore be considered not as a huge improvment in classification accuracy, but as a balanced combination of comparable predictive performance, improved stability and reduced computational cost.

\bibliography{sn-bibliography}% common bib file

\section*{Declarations}

Some journals require declarations to be submitted in a standardised format. Please check the Instructions for Authors of the journal to which you are submitting to see if you need to complete this section. If yes, your manuscript must contain the following sections under the heading `Declarations':

\begin{itemize}
\item Funding: This research received no specific grant from any funding agency in the public, commercial, or not-for-profit sectors.
\item Competing interests: The authors declare that they have no competing interests.
\item Ethics approval and consent to participate: The studies involving human participants were reviewed and approved by the Maria Curie-Skłodowska University Bioethical Commission. The participants provided their written informed consent to participate in this study.
\item Consent for publication: Not applicable. The manuscript does not contain identifiable personal data,
images, or other information requiring individual consent for publication.
\item Data availability: The EEG data analyzed in the present study contain human participant data and
are not publicly available. The data may be made available by the corresponding
author upon reasonable request, subject to applicable ethical and privacy
restrictions.
\item Author contribution: F.P.: Principal Investigator of the project, research project conceptualization at applied neuroinformatics dimension, concept of convolutional neural networks application, data processing, data analysis, manuscript preparation, models design and implementation, EEG recordings; G.M.W.: Key Investigator of the research, research idea and methodological support, manuscript preparation; 
\end{itemize}

\noindent
If any of the sections are not relevant to your manuscript, please include the heading and write `Not applicable' for that section. 

\bigskip
\begin{flushleft}%
Editorial Policies for:

\bigskip\noindent
Springer journals and proceedings: \url{https://www.springer.com/gp/editorial-policies}

\bigskip\noindent
Nature Portfolio journals: \url{https://www.nature.com/nature-research/editorial-policies}

\bigskip\noindent
\textit{Scientific Reports}: \url{https://www.nature.com/srep/journal-policies/editorial-policies}

\bigskip\noindent
BMC journals: \url{https://www.biomedcentral.com/getpublished/editorial-policies}
\end{flushleft}

\end{document}